\documentclass[letterpaper, 10 pt, conference]{ieeeconf}

\usepackage{graphicx}
\usepackage{amsmath}
\usepackage{amssymb}
\usepackage{citesort}
\usepackage{url}
\usepackage{xcolor}
\usepackage{fancyhdr}
\usepackage[hidelinks]{hyperref}
\DeclareMathOperator*{\argmin}{\arg\!\min}

\IEEEoverridecommandlockouts
\title{\LARGE \bf
 Self-supervised Learning of LiDAR Odometry for Robotic Applications}

\author{Julian Nubert$^{1,2}$, Shehryar Khattak$^{1}$ and Marco Hutter$^{1}$
\thanks{*This work is supported in part by the Max Planck ETH Center for Learning Systems, the European Research Council (ERC) under the European Union’s Horizon 2020 research and innovation programme grant agreement No.852044, the Swiss National Science Foundation through the National Centre of Competence in Research Robotics (NCCR) and the Swiss National Science Foundation (SNSF) as part of project No.188596.}
\thanks{$^{1}$The authors are with the Robotic Systems Lab, ETH Z\"urich, {\tt\small\{nubertj, skhattak, mahutter\}@.ethz.ch}.}%
\thanks{$^{2}$The author is with the Max Planck ETH Center for Learning Systems, Germany/Switzerland.}%
}

\begin{document}

\maketitle
\thispagestyle{empty}
\pagestyle{empty}

\begin{abstract}
Reliable robot pose estimation is a key building block of many robot autonomy pipelines, with LiDAR localization being an active research domain. In this work, a versatile self-supervised LiDAR odometry estimation method is presented, in order to enable the efficient utilization of all available LiDAR data while maintaining real-time performance. The proposed approach selectively applies geometric losses during training, being cognizant of the amount of information that can be extracted from scan points. In addition, no labeled or ground-truth data is required, hence making the presented approach suitable for pose estimation in applications where accurate ground-truth is difficult to obtain. Furthermore, the presented network architecture is applicable to a wide range of environments and sensor modalities without requiring any network or loss function adjustments. The proposed approach is thoroughly tested for both indoor and outdoor real-world applications through a variety of experiments using legged, tracked and wheeled robots, demonstrating the suitability of learning-based LiDAR odometry for complex robotic applications.
\end{abstract}


\section{Introduction}\label{sec:intro}
Reliable and accurate pose estimation is one of the core components of most robot autonomy pipelines, as robots rely on their pose information to effectively navigate in their surroundings and to efficiently complete their assigned tasks. 
In the absence of external pose estimates, e.g. provided by GPS or motion-capture systems, robots utilize on-board sensor data for the estimation of their pose. Recently, 3D LiDARs have become a popular choice due to reduction in weight, size, and cost. LiDARs can be effectively used to estimate the 6-DOF robot pose as they provide direct depth measurements, allowing for the estimation at scale while remaining unaffected by certain environmental conditions, such as poor illumination and low-texture.

\begin{figure}[ht]
\centering
    \includegraphics[width=0.99\columnwidth]{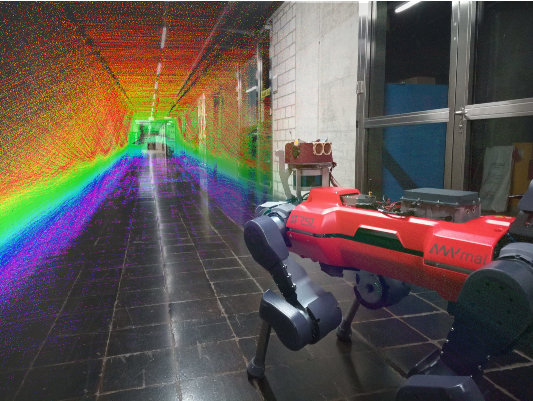}
\caption{ANYmal during an autonomous exploration and mapping mission at ETH Z\"urich, with the height-colored map overlayed on-top of the image. The lack of environmental geometric features as well as rapid rotation changes due to motions of walking robots make the mission challenging.
}\label{fig:main}
\vspace{-0.6cm}
\end{figure}

To estimate the robot's pose from LiDAR data, established model-based techniques such as Iterative Closest Point (ICP)~\cite{GICP,ICPs} typically perform a scan-to-scan alignment between consecutive LiDAR scans. 
However, to maintain real-time operation, in practice only a subset of available scan data is utilized. This subset of points is selected by either down-sampling or by selecting salient scan points deemed to contain the most information~\cite{loam_autonomous_robots}. 
However, such data reduction techniques can lead to a non-uniform spatial distribution of points, as well as to an increase in sensitivity of the underlying estimation process to factors such as the mounting orientation of the sensor.
More complex features~\cite{SPIN,FPFH,SHOT} can be used to make the point selection process invariant to sensor orientation and robot pose, however high-computational cost makes them unsuitable for real-time robot operation. Furthermore, although using all available scan data may not be necessary, yet it has been shown that utilizing more scan data up to a certain extent can improve the quality of the scan-to-scan alignment process~\cite{PCLSampling}.

In order to utilize all available scan data efficiently, learning-based approaches offer a potential solution for the estimation of the robot's pose directly from LiDAR data. Similar approaches have been successfully applied to camera data and have demonstrated promising results~\cite{li2018undeepvo}. However, limited work has been done in the field of learning-based robot pose estimation using LiDAR data, in particular for applications outside the domain of autonomous driving. Furthermore, most of the proposed approaches require labelled or supervision data for their training, making them limited in scope as annotating LiDAR data is particularly time consuming~\cite{semanticKitti}, and obtaining accurate ground-truth data for longer missions, especially indoors, is particularly difficult.

Motivated by the challenges mentioned above, this work presents a self-supervised learning-based approach that utilizes LiDAR data for robot pose estimation. Due to the self-supervised nature of the proposed approach, it  does not require any labeled or ground-truth data during training. In contrast to previous work, arbitrary methods can be utilized for performing the normal computation on the training set; in this work PCA is used. Furthermore, the presented approach does not require expensive pre-processing of the data during inference; instead only data directly available from the LiDAR is utilized. As a result, the proposed approach is computationally lightweight and is capable of operating in real-time on a mobile-class CPU. The performance of the proposed approach is verified and compared against existing methods on driving datasets. Furthermore, the suitability towards complex real-world robotic applications is demonstrated for the first time by conducting autonomous mapping missions with the quadrupedal robot ANYmal~\cite{ANYmal}, shown in operation in Figure~\ref{fig:main}, as well as evaluating the mapping performance on DARPA Subterranean (SubT) Challenge datasets~\cite{DarpaDataset}. Finally, the code of the proposed method is publicly available for the benefit of the robotics community\footnote{\label{note:github} \url{https://github.com/leggedrobotics/DeLORA}}.

\section{Related Work}\label{sec:related}
To estimate robot pose from LiDAR data, traditional or model-based approaches, such as ICP~\cite{GICP,ICPs}, typically minimize either point-to-point or point-to-plane distances between points of consecutive scans. In addition, to maintain real-time performance, these approaches choose to perform such minimization on only a subset of available scan data. Naively, this subset can be selected by sampling points in a random or uniform manner. However, this approach can either fail to maintain uniform spatial scan density or inaccurately represent the underlying local surface structure. As an alternative, works presented in~\cite{NDT} and~\cite{SUMA} aggregate the depth and normal information of local point neighborhoods and replace them by more compact Voxel and Surfel representations, respectively. The use of such representations has shown an improved real-time performance, nevertheless, real scan data needs to be maintained separately as it gets replaced by its approximation. In contrast, approaches such as~\cite{loam_autonomous_robots,legoLOAM}, choose to extract salient points from individual LiDAR scan-lines in order to reduce input data size while utilizing original scan data and maintaining a uniform distribution. These approaches have demonstrated excellent results, yet such scan-line point selection makes these approaches sensitive to the mounting orientation of the sensor, as only depth edges perpendicular to the direction of LiDAR scan can be detected. To select salient points invariant to sensor orientation,~\cite{CLS} proposes to find point pairs across neighboring scan lines. However, such selection comes at increased computational cost, requiring random sampling of a subset of these point pairs for real-time operation.

To efficiently utilize all available scan data without sub-sampling or hand-crafted feature extraction, learning-based approaches can provide a potential solution. In~\cite{FCGF,3DMatch}, the authors demonstrate the feasibility of using learned feature points for LiDAR scan registration. Similarly, for autonomous driving applications,~\cite{dmlo} and~\cite{l3net} deploy supervised learning techniques for scan-to-scan and scan-to-map matching purposes, respectively. However, these approaches use learning as an intermediate feature extraction step, while the estimation is obtained via geometric transformation~\cite{dmlo} and by solving a classification problem~\cite{l3net}, respectively. To estimate robot pose in an end-to-end manner from LiDAR data,~\cite{velas} utilizes Convolution Neural Networks to estimate relative translation between consecutive LiDAR scans, which is then separately combined with relative rotation estimates from an IMU. In contrast,~\cite{lonet} demonstrates the application of learning-based approaches towards full 6-DOF pose estimation directly from LiDAR data alone. However, it should be noted that all these techniques are supervised in nature, and hence rely on the provision of ground-truth supervision data for training.
Furthermore, these techniques are primarily targeted towards autonomous driving applications which, as noted by~\cite{velas}, are very limited in their rotational pose component.

Unsupervised approaches have shown promising results with camera data~\cite{zhou2017unsupervised,li2018undeepvo,zhu2018robustness}. However, the only related work similar to the proposed approach and applied to LiDAR scans is presented in~\cite{deeplo}, which, while performing well for driving use-cases, skips demonstration for more complex robotic applications. Moreover, it requires a simplified normal computation due to its network and loss design, as well as an additional field-of view loss in order to avoid divergence of the predicted transformation.

In this work, a self-supervised learning-based approach is presented that can estimate 6-DOF robot pose directly from consecutive LiDAR scans, while being able to operate in real-time on a mobile-CPU. Furthermore, due to a novel design, arbitrary methods can be used for the normals-computation, without need for explicit regularization during training. Finally, the application of the proposed work is not limited to autonomous driving, and experiments with legged and tracked robots as well as three different sensors demonstrate the variety of real-world applications.

\section{Proposed Approach}\label{sec:approach}
\begin{figure*}[ht!]
\includegraphics[width=\textwidth]{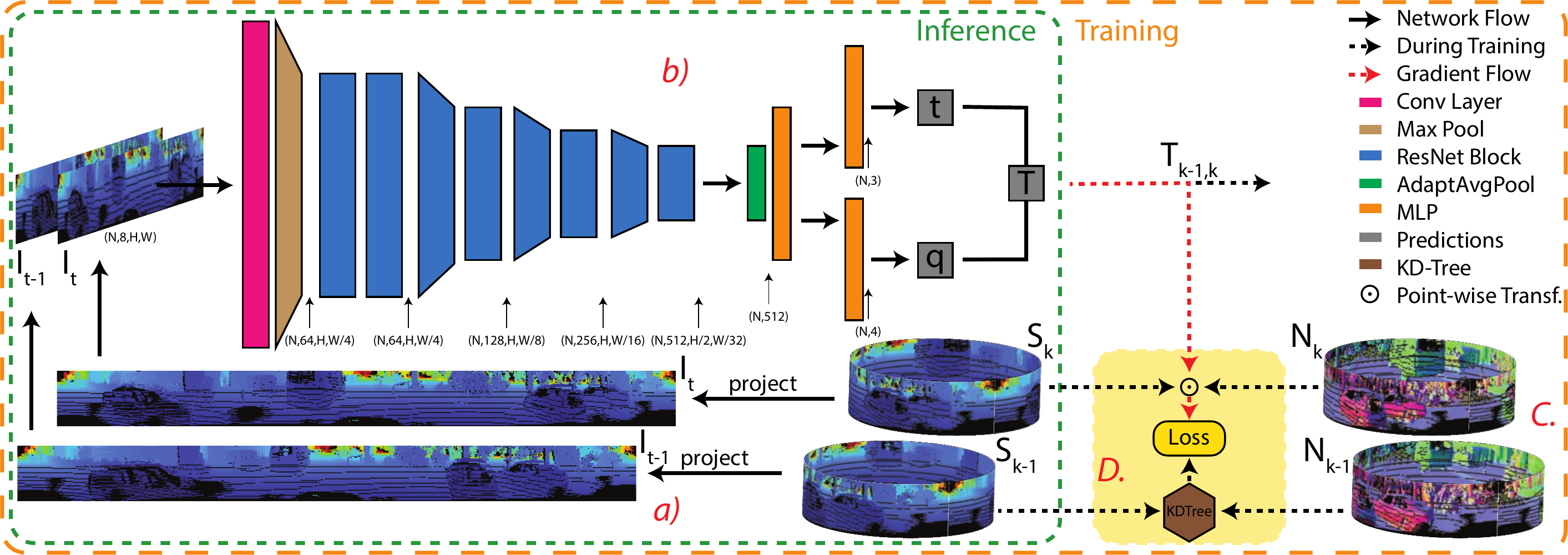}
\caption{Visualization of the proposed approach. The letters \textit{a)}, \textit{b)}, \textit{C.} and \textit{D.} correspond to the identically named subsections in Sec.~\ref{sec:approach}. Starting from the previous and current sensor inputs $\mathcal{S}_{t-1}$ and $\mathcal{S}_{t}$, two LiDAR range images $\mathcal{I}_{t-1}, \mathcal{I}_{t}$ are created which are then fed into the network. The output of the network is a geometric transformation, which is applied to the source scan and normals $\mathcal{S}_{k}, \mathcal{N}_{k}$. After finding target correspondences with the aid of a KD-Tree, a geometric loss is computed, which is then back-propagated to the network during training.}
\label{fig:approach_overview}
\vspace{-0.6cm}
\end{figure*}
In order to cover a large spatial area around the sensor, one common class of LiDARs measures point distances while rotating about its own yaw axis. As a result, a data flow of detected 3D points is generated, often bundled by the sensor as full point cloud scans $\mathcal{S}$.
This work proposes a robot pose estimator which is self-supervised in nature and only requires LiDAR point cloud scans $\mathcal{S}_k,\mathcal{S}_{k-1}$ from the current and previous time steps as its input.

\subsection{Problem Formulation}
\label{sec:problem_formulation}

At every time step $k \in \mathbb{Z^+}$, the aim is to estimate a relative homogeneous transformation $T_{k-1,k} \in SE(3)$, which transforms poses expressed in the sensor frame at time step $k$ into the previous sensor frame at time step $k-1$. As an observation of the world, the current and previous point cloud scans $\mathcal{S}_k \in \mathbb{R}^{n_k \times 3}$ and $\mathcal{S}_{k-1} \in \mathbb{R}^{n_{k-1} \times 3}$ are provided, where $n_k$ and $n_{k-1}$ are the number of point returns in the corresponding scans. Additionally, as a pre-processing step and only for training purposes, normal vectors $\mathcal{N}_k(\mathcal{S}_k)$ are extracted. Due to measurement noise, the non-static nature of environments and the motion of the robot in the environment, the relationship between the transformation $T_{k-1,k}$ and the scans can be described by the following unknown conditional probability density function:
\begin{equation}
p(T_{k-1,k} | \mathcal{S}_{k-1},\mathcal{S}_{k}).    
\end{equation}
In this work, it is assumed that a unique deterministic map  $\mathcal{S}_{k-1},\mathcal{S}_{k} \mapsto T_{k-1,k}$ exists, of which an approximation $\Tilde{T}_{k-1,k}(\theta, \mathcal{S}_{k-1}, \mathcal{S}_k)$ is modeled by a deep neural network. Here, $\theta \in \mathbb{R}^{P}$ denotes the weights and biases of the network, with $P$ being the number of trainable parameters. During training, the values of $\theta$ are obtained by optimizing a geometric loss function $\mathcal{L}$, s.t. $\theta^* = \argmin\limits_{\theta} \mathcal{L}(\Tilde{T}_{k-1,k}(\theta), \mathcal{S}_{k-1}, \mathcal{S}_{k}, \mathcal{N}_{k-1}, \mathcal{N}_{k})$, which will be discussed in more detail in Sec.~\ref{sec:geometric_loss}.

\subsection{Network Architecture and Data Flow}
As this work focuses on general robotic applications, a priority in the approach's design is given to achieve real-time performance on hardware that is commonly deployed on robots. For this purpose, computationally expensive pre-processing operations such as calculation of normal vectors, as e.g. done in~\cite{lonet}, are avoided. Furthermore, during inference the proposed approach only requires raw sensor data for its operation. An overview of the proposed approach is presented in Figure~\ref{fig:approach_overview}, with red letters \emph{a), b), C., D.} providing references to the
following subsections and paragraphs.

\paragraph{Data Representation}
There are three common techniques to perform neural network operations on point cloud data: i) mapping the point cloud to an image representation and applying $2D$-techniques and architectures~\cite{airborneparticle,milioto2019rangenet++}, ii) performing $3D$ convolutions on voxels~\cite{VoxNet,airborneparticle} and iii) to perform operations on disordered point cloud scans~\cite{pointnet,pointnet++}. Due to PointNet's~\cite{pointnet} invariance to rigid transformations and the high memory-requirements of $3D$ voxels for sparse LiDAR scans, this work utilizes the $2D$ image representation of the scan as the input to the network, similar to DeepLO~\cite{deeplo}. 

To obtain the image representation, a geometric mapping of the form $\phi: \mathbb{R}^{n \times 3} \rightarrow \mathbb{R}^{4 \times H \times W}$ is applied, where $H$ and $W$ denote the height and width of the image, respectively. Coordinates $(u,v)$ of the image are calculated by discretizing the azimuth and polar angles in spherical coordinates, while making sure that only the nearest point is kept at each pixel location. A natural choice for $H$ is the number of vertical scan-lines of the sensor, whereas $W$ is typically chosen to be smaller than the amount of points per ring, in order to obtain a dense image (cf. \textit{a)} in Figure.~\ref{fig:approach_overview}). In addition to $3D$ point coordinates, range is also added, yielding $(x,y,z,r)^\intercal$ for each valid pixel of the image, given as $\mathcal{I} = \phi(\mathcal{S})$.

\paragraph{Network}
In order to estimate $\Tilde{T}_{t-1,t}(\theta, \mathcal{I}_{k-1}, \mathcal{I}_k)$, a network architecture consisting of a combination of convolutional, adaptive average pooling, and fully connected layers is deployed, which produces a fixed-size output independent of the input dimensions of the image. For this purpose, $8$ ResNet~\cite{ResNet}-like blocks, which have proven to work well for image to value/label mappings, constitute the core of the architecture. 
In total, the network employs approximately $10^7$ trainable parameters. After generating a feature map of $(N,512,\frac{H}{2},\frac{W}{32})$ dimensions, adaptive average pooling along the height and width of the feature map is performed to obtain a single value for each channel. The resulting feature vector is then fed into a single multi-layer perceptron (MLP), before splitting into two separate MLPs for predicting translation $t \in \mathbb{R}^3$ and rotation in the form of a quaternion $q \in \mathbb{R}^4$. Throughout all convolutional layers, circular padding is applied, in order to achieve the same behavior as for a true (imaginary) $360$° circular image. After normalizing the quaternion, $\bar{q} = \frac{q}{|q|}$, the transformation matrix $\Tilde{T}_{k-1,k}(\bar{q}(\theta,\mathcal{S}_{k-1}, \mathcal{S}_k), t(\theta,\mathcal{S}_{k-1}, \mathcal{S}_k))$ is computed.

\subsection{Normals Computation}
\label{sec:normals_computation}
Learning rotation and translation at once is a difficult task~\cite{velas}, since both impact the resulting loss independently and can potentially make the training unstable. However, recent works~\cite{lonet,deeplo} that have utilized normal vector estimates in their loss functions have demonstrated good estimation performance. Nevertheless, utilizing  normal vectors for loss calculation is not trivial, and due to difficult integration of \textit{"direct optimization approaches into the learning process"}~\cite{deeplo}, DeepLO computes its normal estimates with simple averaging methods by explicitly computing the cross product of vertex-points in the image. In the proposed approach, no loss-gradient needs to be back-propagated through the normal vector calculation (i.e. the eigen-decomposition), as normal vectors are calculated in advance. Instead, normal vectors computed offline are simply rotated using the rotational part of the computed transformation matrix, allowing for simple and fast gradient flow with arbitrary normal computation methods. Hence, in this work normal estimates are computed via a direct optimization method, namely principal component analysis (PCA) of the estimated covariance matrix of neighborhoods of points as described in~\cite{serafin2016fast}, allowing for more accurate normal vector predictions.  
Furthermore, normals are only computed for points that have a minimum number of valid neighbors, where the validity of neighbors is dependent on their depth difference from the point of interest $x_i$, i.e. $|\text{range}(x_i) - \text{range}(x_{nb})|_2 \leq \alpha$, with $\alpha$ empirically set to $0.5 m$ in the conducted experiments.

\vspace{-0.5em}
\subsection{Geometric Loss}
\label{sec:geometric_loss}
In this work, a combination of geometric losses akin to the cost functions in model-based methods~\cite{ICPs} are used, namely point-to-plane and plane-to-plane loss. 
The rigid body transformation $\Tilde{T}_{k-1,k}$ is applied to the source scan, s.t. $\Tilde{\mathcal{S}}_{k-1} = \Tilde{T}_{k-1,k} \odot \mathcal{S}_{k}$, and its rotational part to all source normal vectors, s.t. $\Tilde{\mathcal{N}}_{k-1} = \text{rot}(\Tilde{T}_{k-1,k}) \odot \mathcal{N}_{k}$, where $\odot$ denotes an element-wise matrix multiplication. The loss function then incentivizes the network to generate a $\Tilde{T}_{k-1,k}$, s.t. $\Tilde{\mathcal{S}}_{k-1}, \Tilde{\mathcal{N}}_{k-1}$ match $\mathcal{S}_{k-1}, \mathcal{N}_{k-1}$ as close as possible. 

\paragraph{Correspondence Search}
In contrast to~\cite{deeplo,lonet} where image pixel locations are used as correspondences, this work utilizes a full correspondence search in $3D$ using a KD-Tree~\cite{kd-tree} among the transformed source and target. This has two main advantages: First, as opposed to~\cite{deeplo}, there is no need for an additional field-of-view loss, since correspondences are also found for points that are mapped to regions outside of the image boundaries. Second, this allows for the handling of cases close to sharp edges, which, when using discretized pixel locations only~\cite{deeplo}, can lead to wrong correspondences for points with large depth deviations. Once point correspondences have been established, the following two loss functions can be computed.

\paragraph{Point-to-Plane Loss}
For each point $\hat{s}_b$ in the transformed source scan $\hat{\mathcal{S}}_{k-1}$, the distance to the associated point $s_b$ in the target scan is computed, and projected onto the target surface at that position, i.e.
\begin{equation}
    \mathcal{L}_{\text{p2n}} = \frac{1}{n_k} \sum_{b=1}^{n_k} |(\hat{s}_b - s_b) \cdot n_b|_2^2,
\end{equation}
where $n_b$ is the target normal vector. If no normal exists either at the source or at the target point, the point is considered invalid and omitted from the loss calculation.

\paragraph{Plane-to-Plane Loss}
In the second loss term, the surface orientation around the two points is compared. Let $\hat{n}_b$ and $n_b$ be the normal vectors at the transformed source and target locations, then the loss is computed as follows:
\begin{equation}
    \mathcal{L}_{n2n} = \frac{1}{n_k} \sum_{b=1}^{n_k} |\hat{n}_b - n_b|_2^2.
\end{equation}
Again, point correspondences are only selected for the loss computation if normals are present at both point locations. 

The final loss is then computed as $\mathcal{L} = \lambda \cdot \mathcal{L}_{p2n} + \mathcal{L}_{n2n}$. The ratio $\lambda$ did not significantly impact the performance, with both terms $\mathcal{L}_{p2n}$ and $\mathcal{L}_{n2n}$ converging independently. For the conducted experiments $\lambda$ was set to $1$.

\section{Experimental Results}\label{sec:evaluation}
To thoroughly evaluate the proposed approach, testing is performed on three robotic datasets using different robot types, different LiDAR sensors and sensor mounting orientations. First, using the quadrupedal robot ANYmal, the suitability of the proposed approach for real-world autonomous missions is demonstrated by integrating its pose estimates into a mapping pipeline and comparing against a state-of-the-art model-based approach~\cite{loam_autonomous_robots}. Next, reliability of the proposed approach is demonstrated by applying it to datasets from the DARPA SubT Challenge~\cite{DarpaDataset}, collected using a tracked robot, and comparing the built map against the ground-truth map. Finally, to aid numerical comparison with existing work, an evaluation is conducted on the KITTI odometry benchmark~\cite{Geiger2012CVPR}.

The proposed approach is implemented using \textit{PyTorch}~\cite{pytorch}, utilizing the KD-Tree search component from \textit{SciPy}. For testing, the model is embedded into a ROS~\cite{ROS} node. The full implementation is made publicly available\textsuperscript{\ref{note:github}}.

\begin{figure*}[t!]
\centering
    \includegraphics[width=\textwidth]{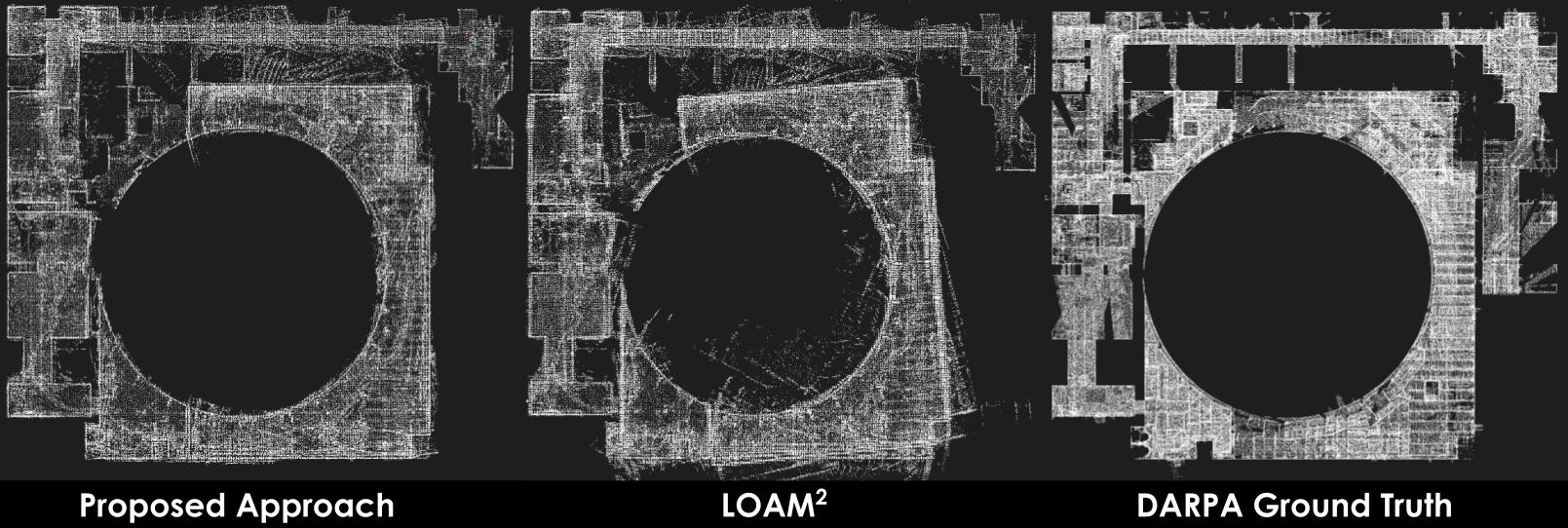}
\caption{Comparison of maps created by using pose estimates from the proposed approach and LOAM\textsuperscript{$2$} implementation against ground-truth map, as provided in the DARPA SubT Urban Circuit dataset. More consistent mapping results can be noted when comparing the proposed map with the ground-truth.}
\label{fig:darpa}
\vspace{-0.6cm}
\end{figure*}

\subsection{ANYmal: Autonomous Exploration Mission}\label{sec:anymal}
To demonstrate the suitability for complex real-world applications, the proposed approach is tested on data collected during autonomous exploration and mapping missions conducted with the ANYmal quadrupedal robot~\cite{ANYmal}. In contrast to wheeled robots and cars, ANYmal with its learning-based controller~\cite{LeeController} has more variability in roll and pitch angles during walking. Additionally, rapid large changes in yaw are introduced due to the robot's ability to turn on spot. During these experiments, the robot was tasked to autonomously explore~\cite{GBPlanner} and map~\cite{CompSLAM_2020} a previously unknown indoor environment and autonomously return to its start position. The experiments were conducted in the basement of the CLA building at ETH Z\"urich, containing long tunnel-like corridors, as shown in Figure~\ref{fig:main}, and during each mission ANYmal traversed an average distance of $250$ meters. 

For these missions ANYmal was equipped with a Velodyne VLP-16 Puck Lite LiDAR. In order to demonstrate the robustness of the proposed method, during the test mission the LiDAR was mounted in upside-down orientation, while during training it was mounted in the normal upright orientation. To record the training set, two missions were conducted with the robot starting from the right side-entrance of the main course. For testing, the robot started its mission from the previously unseen left side-entrance, located on the opposing end of the main course. 

\begin{figure}[ht!]
\centering
    \includegraphics[width=0.99\columnwidth]{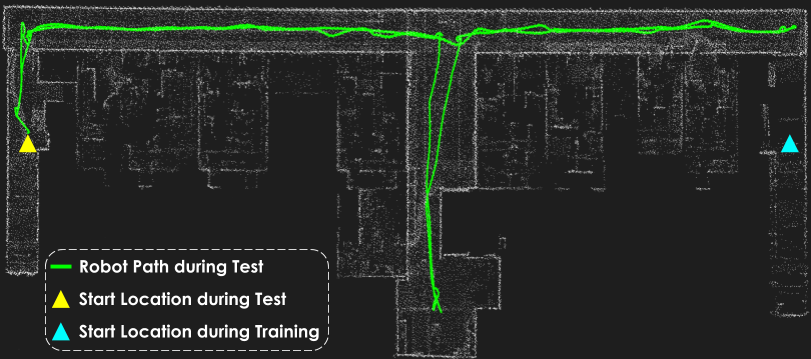}
\caption{Map created for an autonomous test mission of ANYmal robot. The robot path during the mission is shown in green, with the triangles highlighting the different starting positions for the training and test sets.}
\label{fig:anymal_map}
\vspace{-0.2cm}
\end{figure}

During training and test missions, the robot never visited the starting locations of the other mission as they were physically closed off. To demonstrate the utility of the proposed method for mapping applications, the estimated robot poses were combined with the mapping module of LOAM~\cite{loam_autonomous_robots}. Figure~\ref{fig:anymal_map} shows the created map, the robot path during test, as well as the starting locations for training and test missions. During testing, a single prediction takes about $48$ms on an \textit{i7-8565U} low-power laptop CPU, and $13$ms on a small \textit{GeForce MX250} laptop GPU, with $n_k \approx 32,000$, $H=16$, $W=720$.

\begin{figure}[ht!]
    \centering
         \centering
         \includegraphics[width=0.49\columnwidth]{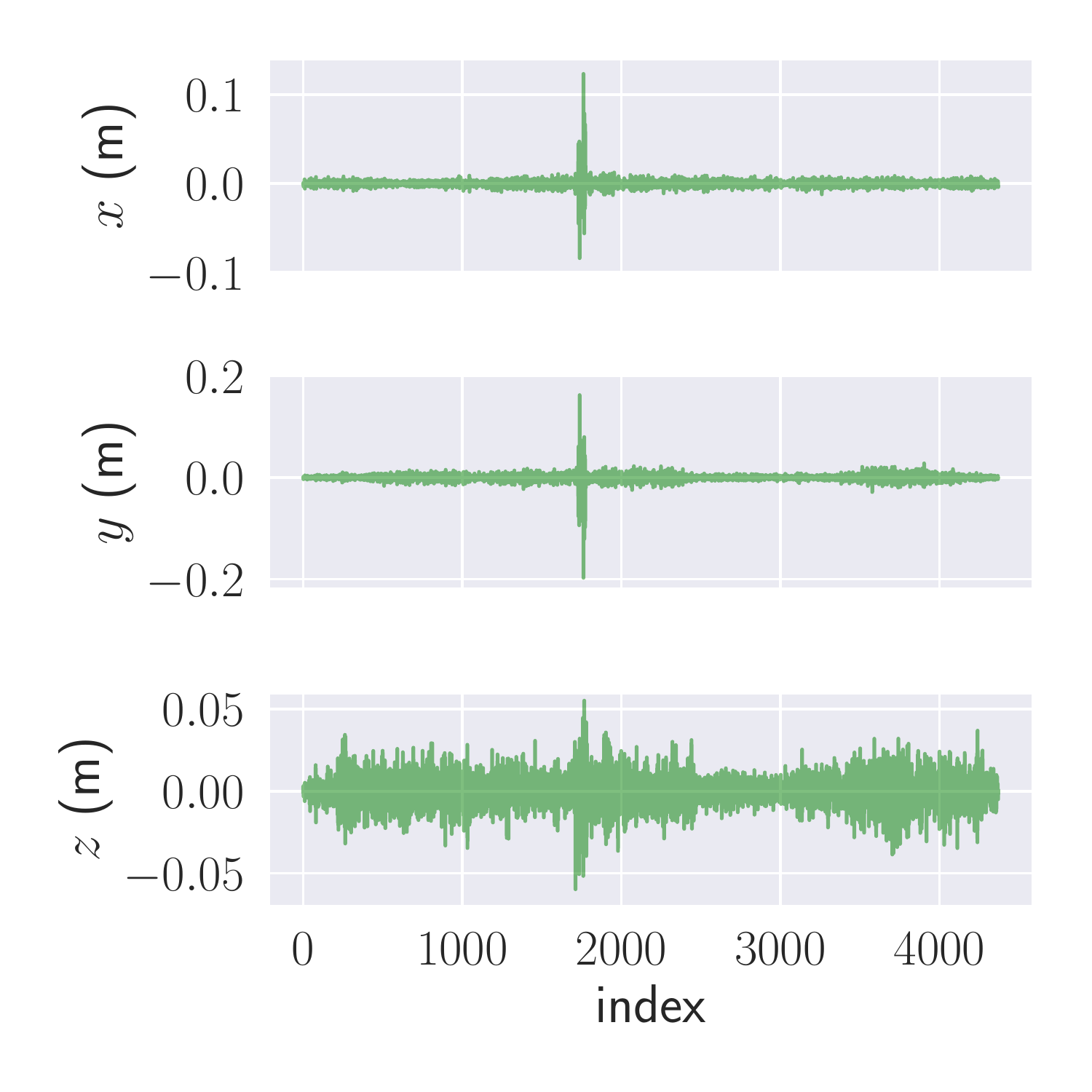}
         \centering
         \includegraphics[width=0.49\columnwidth]{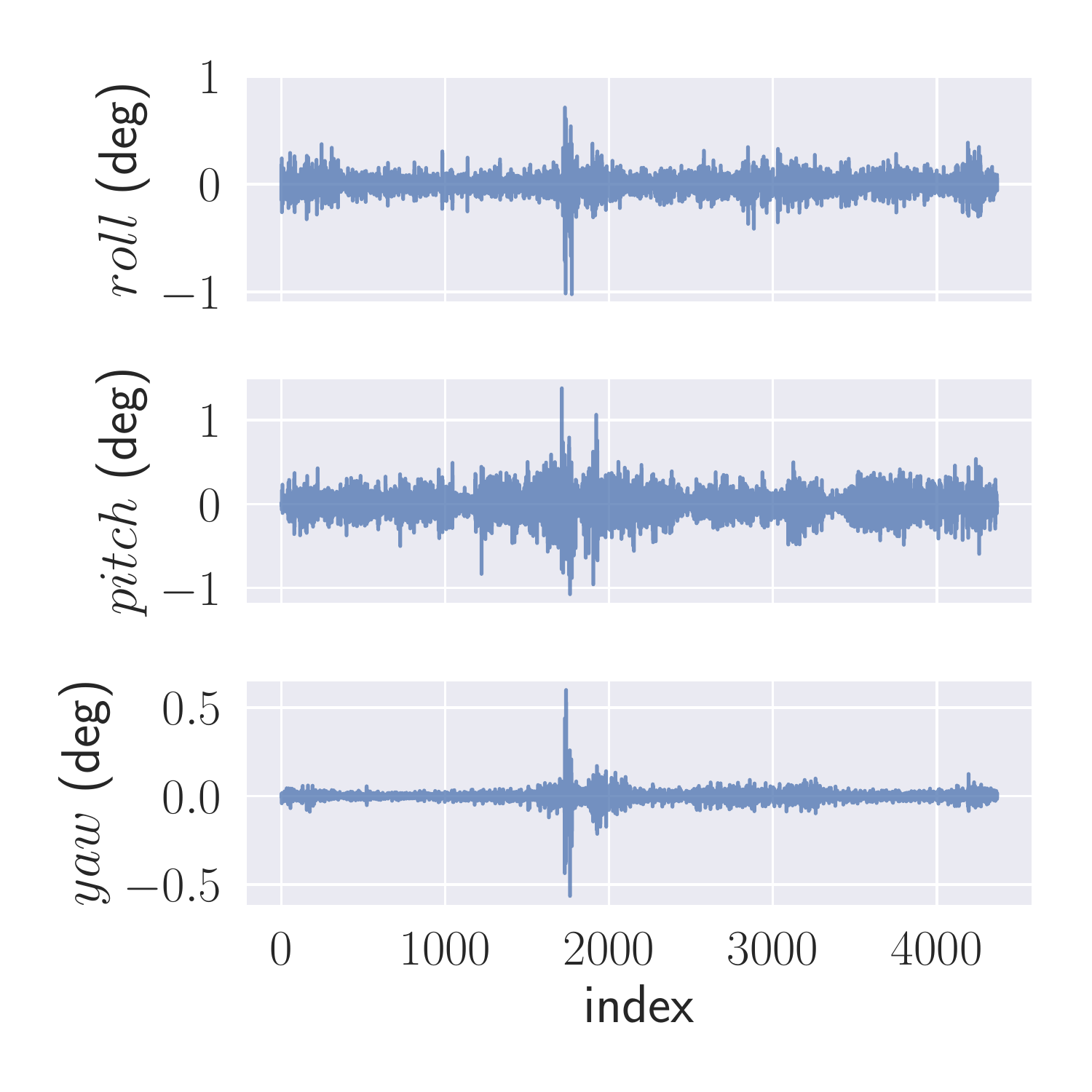}
\caption{Relative translation and rotation deviation plots for each axis between the proposed approach with mapping and LOAM\textsuperscript{\ref{note:loam}} implementation.}
\label{fig:anymal_plot}
\end{figure}

Upon visual inspection it can be noted that the created map is consistent with the environmental layout. Moreover, to facilitate a quantitative evaluation due to absence of external ground-truth, the relative pose estimates of the proposed methods are compared against those provided by a popular open-source LOAM~\cite{loam_autonomous_robots} implementation\footnote{\label{note:loam}\url{https://github.com/laboshinl/loam_velodyne}}. 
The quantitative results are presented in Table~\ref{tab_anymal}, with corresponding error plots shown in Figure~\ref{fig:anymal_plot}. A very low difference can be observed between the pose estimates produced by the proposed approach and those provided by LOAM, hence demonstrating its suitability for real-world mapping applications.

\begin{table}[ht!]
    \begin{center}
    \caption{Relative pose deviations of the proposed approach with mapping compared to LOAM\textsuperscript{2}, for the ANYMAL dataset.
    }
    \begin{tabular}{c c c c c c c} 
         \hline
          & \multicolumn{6}{c}{Segment length}  \\ [0.5ex] 
          & 5 & 10 & 25 & 40 & 60 & 100 \\ 
          \hline\hline
          $t_{\text{rel}}[\%]$ & $0.345$ & $0.212$ & $0.151$ & $0.160$ & $0.178$ & $0.128$ \\
          $r_{\text{rel}}[\frac{\text{deg}}{\text{10m}}]$  & $0.484$ & $0.274$ & $0.150$ & $0.103$ & $0.069$ & $0.046$
    \end{tabular}
    \label{tab_anymal}
    \end{center}
    \vspace{-0.3cm}
\end{table}

\begin{figure*}[t!]
    \centering
         \centering
         \includegraphics[width=0.24\textwidth]{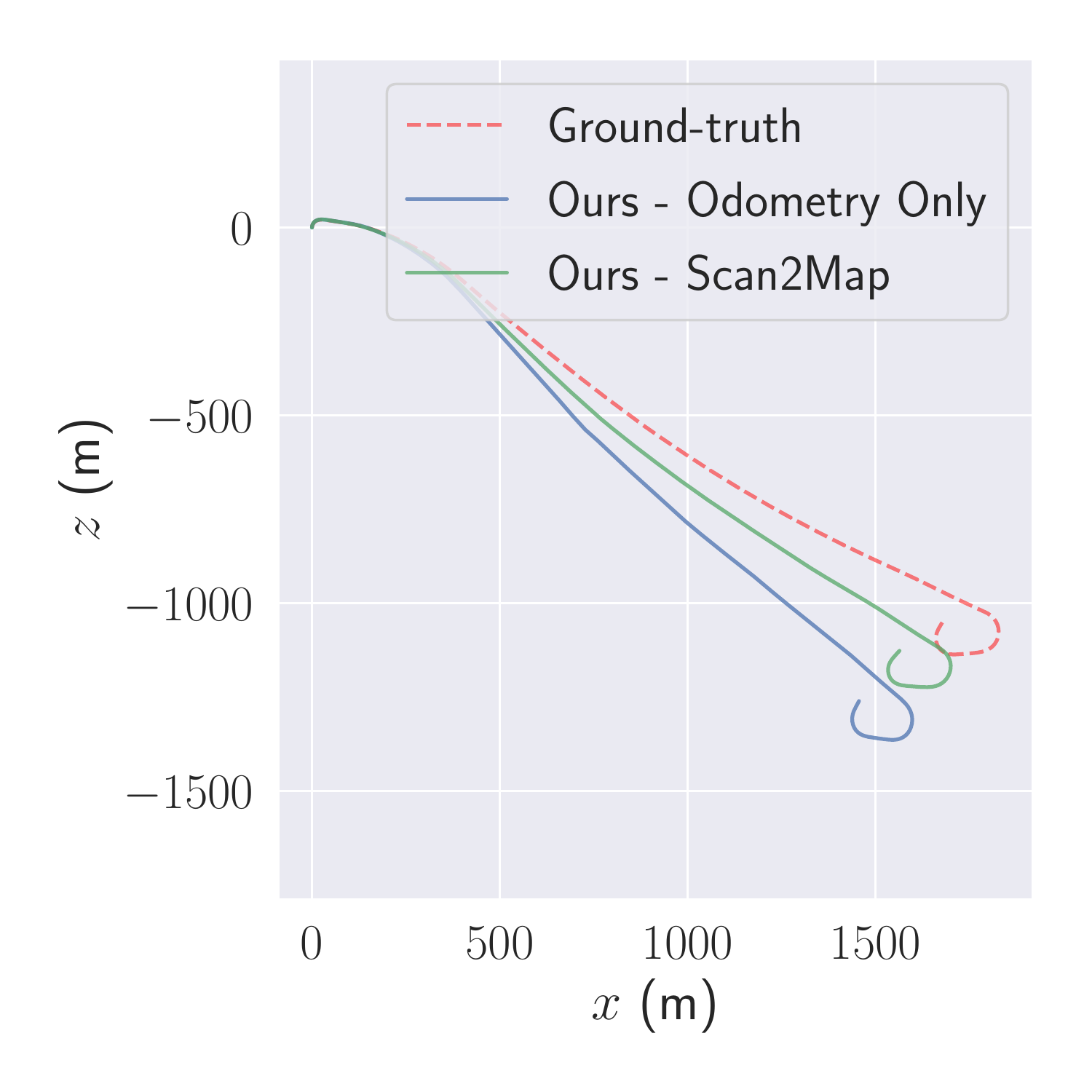}
         \centering
         \includegraphics[width=0.24\textwidth]{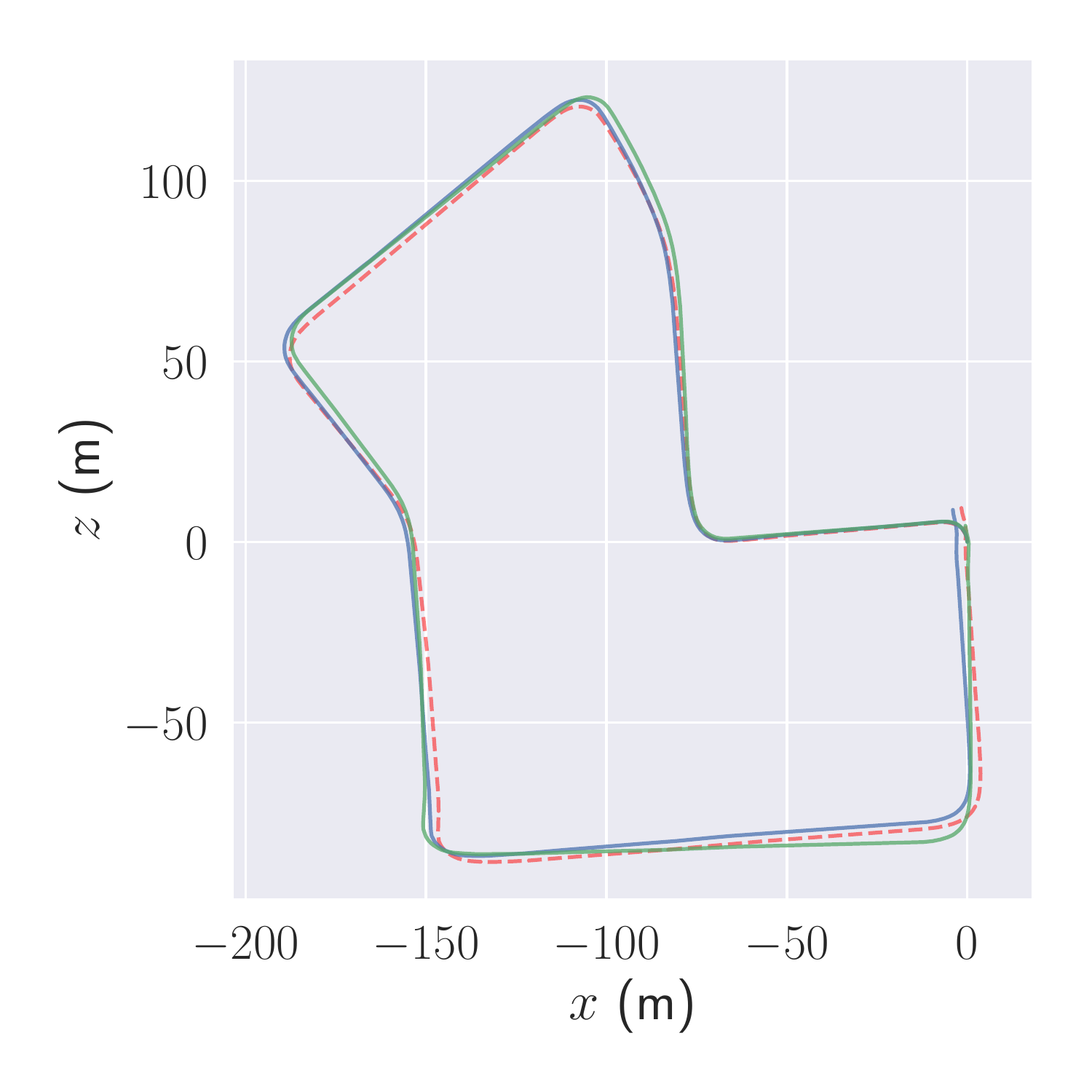}
         \centering
         \includegraphics[width=0.24\textwidth]{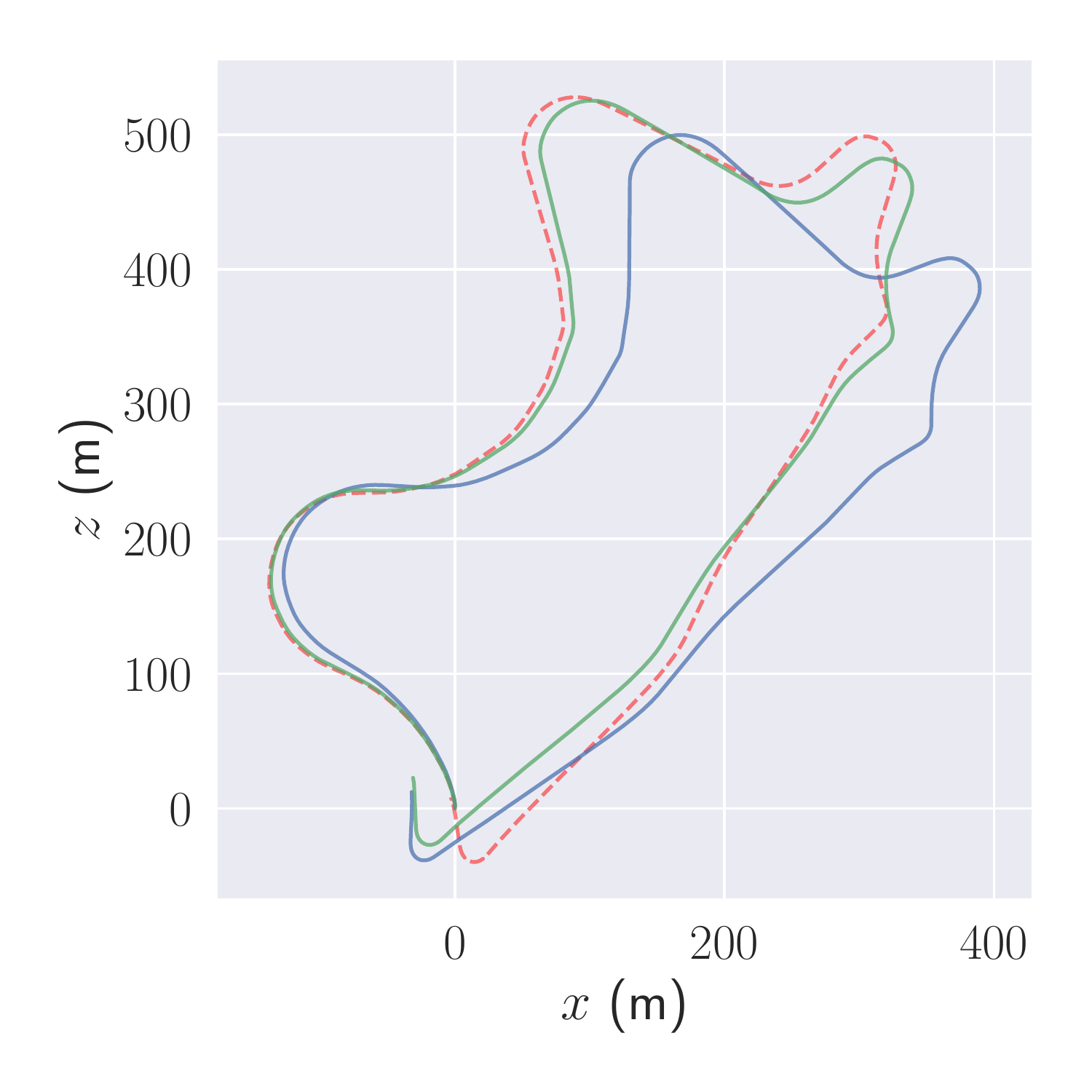}
         \centering
         \includegraphics[width=0.24\textwidth]{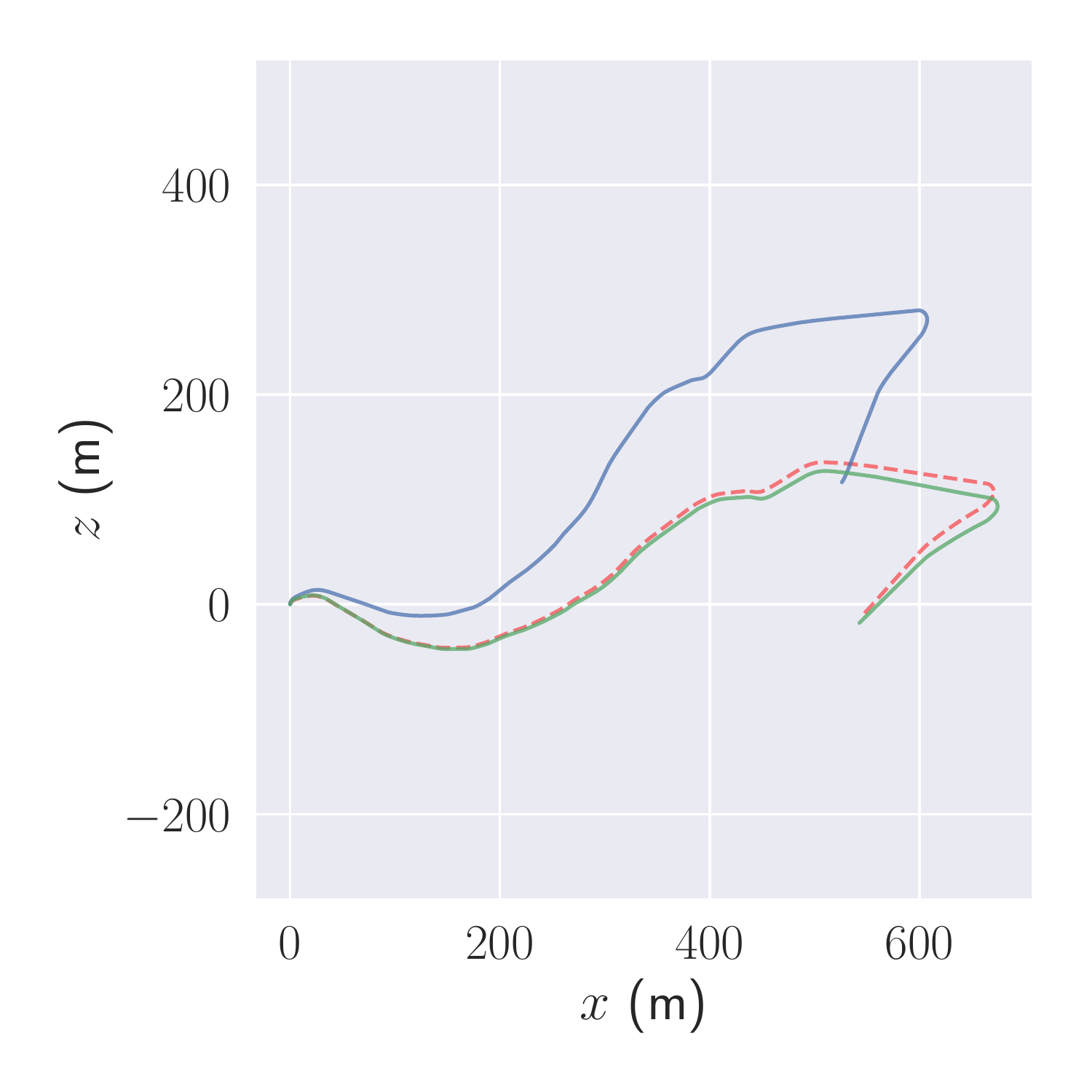}
\caption{Qualitative results of the proposed odometry, as well as the scan-to-map refined version of it. From left to right the following sequences are shown: \texttt{01}, \texttt{07} (training set), \texttt{09}, \texttt{10} (validation set).}
\label{fig:kitti_plot}
\vspace{-0.6cm}
\end{figure*}

\subsection{DARPA SubT Challenge Urban Circuit}
\label{sec:darpa}
Next, the proposed approach is tested on the DARPA SubT Challenge Urban Circuit datasets~\cite{DarpaDataset}. These datasets were collected using an iRobot PackBot Explorer tracked robot carrying an Ouster OS1-64 LiDAR at Satsop Business Park in Washington, USA. The dataset divides the scans of the nuclear power plant facility into Alpha and Beta courses with further partition into upper and lower floors, with a map of each floor provided as ground-truth. It is worth noticing that again a different LiDAR sensor is deployed in this dataset. To test the approach's operational generalization, training was performed on scans from the Alpha course, with testing being done on the Beta course. Similar to before, the robot pose estimates were combined with the LOAM mapping module. The created map is compared with the LOAM implementation\textsuperscript{\ref{note:loam}} and ground-truth maps in Figure~\ref{fig:darpa}. Due to the complex and narrow nature of the environment as well as the ability of the ground robot to make fast in-spot yaw rotations, it can be noted that the LOAM map becomes inconsistent. In contrast, the proposed approach is not only able to generalize and operate in the new environment of the test set but it also provides more reliable pose estimates and produces a more consistent map when compared to the DARPA provided ground-truth map.

\subsection{KITTI: Odometry Benchmark}
To demonstrate real-world performance quantitatively and to aid the comparison to existing work, the proposed approach is evaluated on the KITTI odometry benchmark dataset~\cite{Geiger2012CVPR}. The dataset is split in a training (Sequences \texttt{00-08}) and a test set (Sequences \texttt{09,10}), as also done in DeepLO~\cite{deeplo} and most other learning-based works. The results of the proposed approach are presented in Table~\ref{table:evaluation_00to08}, and are compared to model-based approaches~\cite{loam_autonomous_robots,SUMA}, supervised LiDAR odometry approaches~\cite{lonet,velas} and unsupervised visual odometry methods~\cite{zhu2018robustness,li2018undeepvo,zhou2017unsupervised}. 
\begin{table}[h!]
        \centering
        \caption{Comparison of translational ($[\%]$) and rotational ($[\frac{deg}{100m}]$) errors on all possible sequences of lengths of $\{100,200,\dots,800\}$ meters for the KITTI odometry benchmark.
        }
        \begin{tabular}{c c c c c c c} 
             \hline
              & \multicolumn{2}{c}{Training \texttt{00-08}} & \multicolumn{2}{c}{Sequence \texttt{09}} & \multicolumn{2}{c}{Sequence \texttt{10}} \\
              & 
              $t_{\text{rel}}$ & $r_{\text{rel}}$ & $t_{\text{rel}}$ & $r_{\text{rel}}$ & $t_{\text{rel}}$ & $r_{\text{rel}}$  \\ 
              \hline
              \textbf{Ours} & $3.00$ & $1.38$ & $6.05$ & $2.15$ & $6.44$ & $3.00$ \\
              \textbf{Ours+Map} & $1.78$ & $0.73$ & $1.54$ & $0.68$ & $1.78$ & $0.69$ \\
              DeepLO~\cite{deeplo} & $3.68$ & $0.87$ & $4.87$ & $1.95$ & $5.02$ & $1.83$ \\  
              LO-Net~\cite{lonet} & $1.27$ & $0.67$ & $1.37$ & $0.58$ & $1.80$ & $0.93$ \\
              Velas \textit{et al.}~\cite{velas} & $2.94$
              & NA & $4.94$ & NA & $3.27$ & NA \\
             UnDeepVO~\cite{li2018undeepvo} & $4.54$ & $2.55$ & $7.01$ & $3.61$ & $10.63$ & $4.65$ \\
             SfMLearner~\cite{zhou2017unsupervised} & $28.52$ & $4.67$ & $18.77$ & $3.21$ & $14.33$ & $3.30$ \\
             Zhu \textit{et al.}~\cite{zhu2018robustness} & $5.72$ & $2.35$ & $8.84$ & $2.92$ & $6.65$ & $3.89$ \\
             \begin{tabular}{@{}c@{}} LO-Net+Map\end{tabular} & $0.81$ & $0.44$ & $0.77$ & $0.38$ & $0.92$ & $0.41$ \\
             SUMA~\cite{SUMA} & $3.06$ & $0.89$ & $1.90$ & $0.80$ & $1.80$ & $1.00$ \\
             LOAM~\cite{loam_autonomous_robots} & $1.26$ & $0.50$ & $1.20$ & $0.48$ & $1.51$ & $0.57$ \\
             \hline
        \end{tabular}
        \label{table:evaluation_00to08}
    \vspace{-0.2cm}
\end{table}
Only the \texttt{00-08} mean of the numeric results of LO-Net and Velas \textit{et al.} needed to be adapted, since both were only trained on \texttt{00-06}, yet the results remain very similar to the originally reported ones. 
Results are presented for both, the pure proposed LiDAR scan-to-scan method, as well as for the version that is combined with a LOAM~\cite{loam_autonomous_robots} mapping module, as also used in Section~\ref{sec:anymal} and Section~\ref{sec:darpa}. Qualitative results of the trajectories generated by the predicted odometry estimates, as well as by the map-refined ones are shown in Figure~\ref{fig:kitti_plot}. The proposed approach provides good estimates with little drift, even on challenging sequences with dynamic objects (\texttt{01}), and previously unobserved sequences during training (\texttt{09,10}). Nevertheless, especially for the test set the scan-to-map refinement helps to achieve even better and more consistent results. Quantitatively, the proposed method achieves similar results to the only other self-supervised LiDAR odometry approach~\cite{deeplo}, and outperforms it when combined with mapping, while also outperforming all other unsupervised visual odometry methods~\cite{li2018undeepvo,zhou2017unsupervised,zhu2018robustness}. Similarly, by integrating the scan-to-map refinement, results close to the overall state of the art~\cite{lonet,SUMA,loam_autonomous_robots} are achieved.

Furthermore, to understand the benefit of utilizing both geometric losses, two networks were trained from scratch on a different training/test split of the KITTI dataset. The results are presented in Table~\ref{table:ablation} and demonstrate the benefit of combining plane-to-plane (pl2pl) loss and point-to-plane (p2pl) loss over using the latter one alone, as done in~\cite{deeplo}.
\begin{table}[h!]
    \centering
    \caption{Ablation study showing the translational ($[\%]$) and rotational ($[\frac{deg}{100m}]$) errors for the KITTI benchmark.}
    \begin{tabular}{c c c c c} 
         \hline
          & \multicolumn{2}{c}{Training \texttt{00-06}} & \multicolumn{2}{c}{Test \texttt{07-10}} \\ 
          & $t_{\text{rel}}$ & $r_{\text{rel}}$ & $t_{\text{rel}}$ & $r_{\text{rel}}$ \\ 
          \hline
          p2pl $+$ pl2pl & $3.41$ & $1.44$ & $8.30$ & $3.45$ \\
          p2pl & $6.47$ & $2.72$ & $8.90$ & $4.00$ \\
          \hline
    \end{tabular}
    \label{table:ablation}
    \vspace{-0.2cm}
\end{table}

\section{Conclusions}\label{sec:conclusions}
This work presented a self-supervised learning-based approach for robot pose estimation directly from LiDAR data. The proposed approach does not require any ground-truth or labeled data during training and selectively applies geometric losses to learn domain-specific features while exploiting all available scan information. The versatility and suitability of the proposed approach towards real-world robotic applications is demonstrated by experiments conducted using legged, tracked and wheeled robots operating in a variety of indoor and outdoor environments.
In future, integration of multi-modal sensory information, such as IMU data, will be explored to improve the quality of the estimation process. Furthermore, incorporating temporal components into the network design can potentially make the estimation process robust against local disturbances, which can especially be beneficial for robots traversing over rougher terrains.

\section*{ACKNOWLEDGMENT}
The authors are thankful to Marco Tranzatto, Samuel Zimmermann and Timon Homberger for their assistance with ANYmal experiments.


\bibliographystyle{IEEEtran}
\bibliography{IEEEabrv,ICRA2021}

\end{document}